# Multi-class Skin Cancer Classification Architecture Based on Deep Convolutional Neural Network

Mst Shapna Akter[*], Hossain Shahriar[†], Sweta Sneha[‡], Alfredo Cuzzocrea[§]

[*]Department of Computer Science, Kennesaw State University, USA
[†] Department Information Technology, Kennsaw State University, USA
[‡] Kennesaw State University, USA
[§] iDEA Lab, University of Calabria, Rende, Italy
{[*] makter2@students.kennesaw.edu, [†] hshahria@kennesaw.edu, [‡] ssneha@kennesaw.edu, [§] alfredo.cuzzocrea@unical.it }

*Abstract*—Skin cancer is a deadly disease. Melanoma is a type of skin cancer responsible for the high mortality rate. Early detection of skin cancer can enable patients to treat the disease and minimize the death rate. Skin cancer detection is challenging since different types of skin lesions share high similarities. This paper proposes a computer-based deep learning approach that will accurately identify different kinds of skin lesions. Deep learning approaches can detect skin cancer very accurately since the models learn each pixel of an image. Sometimes humans can get confused by the similarities of the skin lesions, which we can minimize by involving the machine. However, not all deep learning approaches can give better predictions. Some deep learning models have limitations, leading the model to a false-positive result. We have introduced several deep learning models to classify skin lesions to distinguish skin cancer from different types of skin lesions. Before classifying the skin lesions, data preprocessing and data augmentation methods are used. Finally, a Convolutional Neural Network (CNN) model and six transfer learning models such as Resnet-50, VGG-16, Densenet, Mobilenet, Inceptionv3, and Xception are applied to the publi-cally available benchmark HAM10000 dataset to classify seven classes of skin lesions and to conduct a comparative analysis. The models will detect skin cancer by differentiating the cancerous cell from the non-cancerous ones. The models' performance is measured using performance metrics such as precision, recall, f1 score, and accuracy. We receive accuracy of 90, 88, 88, 87, 82, and 77 percent for inceptionv3, Xception, Densenet, Mobilenet, Resnet, CNN, and VGG16, respectively. Furthermore, we develop five different stacking models such as inceptionv3-inceptionv3, Densenet-mobilenet, inceptionv3-Xception, Resnet50-Vgg16, and stack-six for classifying the skin lesions and found that the stacking models perform poorly. We achieve the highest accuracy of 78 percent among all the stacking models.

## I. INTRODUCTION

The superficial layer of skin, called the epidermis, consists of Squamous, Basal, and melanocyte cells. Squamous is the outermost layer of cells. Basal cells are the epidermis' lowermost cells. Melanocyte cells protect deeper layers of skin from sun exposure by producing melanin, a brown pigment substance. Due to ultraviolet light exposure, the DNA mutations induce the growth of skin cells, leading to skin cancer [1]. Melanoma, Squamous Cell Carcinoma, and Basal Cell Carcinoma are the substantial group of skin cancer associated with Squamous, Basal, and Melanocytes cells. Worldwide, Almost 10 million skin cancer deaths took place in 2020. According to the world health organization, it is estimated that, globally, one-third of all diagnosed cancer cases are skin cancer. Nowadays, skin cancer is a global public health issue that causes approximately 5.4 million newly identified skin cancer incidences each year in the United States [2, 3]. However, melanoma alone causes three-fourths of all skin cancer-related deaths, about 10,000 deaths each year in the United States. In Europe, over 1,00,000 cases, and in Australia, nearly 15,229 new cases of melanoma have been accounted for annually [4]. Moreover, an increasing trend of skin cancer has been recorded in the past decades. In the United Kingdom, the percentage of melanoma has increased by 119 percent since the 1990s; in the same duration, it has increased by 250 percent in the United States [5? ]. Skin cancer is an alarming issue, and it should be detected as early as possible. The ritual diagnostic method of detecting skin cancer disease usually is the biopsy method. The biopsy method requires removing a portion of tissue from the cell of the patient's body so that they can analyze it in the laboratory [6]. The whole procedure is painful, time-consuming, and costly. Sometimes, patients may get into trouble due to the hassle of visiting the hospital.

Recently, the most popular non-surgical instruments used for diagnosis systems are macroscopic and dermoscopic images [7]. The macroscopic image has a lower resolution problem since the images are usually captured with a camera and mobile phone [8]. Dermoscopy images are high-resolution skin images derived from visualizing the deeper skin structures [9]. Since multiple skin cancer types share similar symptoms, it becomes challenging for dermatologists to diagnose even with dermoscopy images. Expert dermatologists are limited in their studies and experiences, and it is not possible for a human to recognize all possible appearances of skin cancer. It is found that the ability of dermatologists for skin cancer diagnosis is an average accuracy of 62 percent to 80 percent [10–12] . The accuracy varies with the experience of dermatologists. The worst observation is that the performance further can be dropped for less experienced dermatologists [11]. Such conditions of a skin cancer diagnosis can be very dangerous for cancer patients with false-negative results. It will not improve the deadly condition of the world.

Nowadays, technology has become so advanced that it plays a significant role in the medical sector. The research commu-nity has made significant progress in developing computer-aided diagnosis tools to overcome the life-threatening issue [8, 13, 14]. Modern technology has brought us the concept of a neural network that can perform magnificently for clas-sifying images in the medical domain. However, previous investigations fail to extend their study for multiple classes in skin cancer classification [12, 15–18]. Additionally, those are limited by exploring a few deep learning models [19? , 20]. The model must learn pixel by pixel properly of an image to detect the cancerous cell accurately. There are some limitations in each model, which prevent those models from giving accurate results. Therefore, we can not be sure which model will work best. Previously, some models have performed very well in the medical sector. Since the underlying concept of the deep learning model is like a black box, we cannot even say which model will work best on which kind of dataset. Therefore, we have made a comparative analysis by training several neural network models such as Mobilenet, Inceptionv3, Resnet-50, Vgg-16, Xception, Densenet, and a Convolutional Neural Network (CNN).

The deep learning models are capable of learning seven types of skin cancer such as melanoma, melanocytic nevus, basal cell carcinoma, actinic keratosis, benign keratosis, dermatofibroma, vascular lesion, and Squamous cell carcinoma, and distinguish the cancer cells from non-cancerous cells. At first, we preprocess the data by resizing it to 120X120 resolution. Then we augmented the dataset using augmentation methods such as rotating, horizontal and vertical flipping, std normalization, random zooming, width height shifting, and ZCA whitening. Finally, we fed the images to the neural network models. Furthermore, we developed five stacking ensemble models such as inceptionv3-inceptionv3, Densenet-mobile net, inceptionv3-Xception, Resnet50-Vgg16 and stack-six model and fed the images to the models to check how the stacking ensemble model performs on the skin cancer dataset. We achieve 90 percent accuracy using inceptionv3, which is the highest accuracy among all the models, including the stacking ensemble models. Our proposed method outperformed expert dermatologists, which will contribute to the medical sector by saving many lives. Our comparative analysis has been done using pre-trained deep learning models, which are more efficient than simple models. We have proposed the stacking ensemble models using the weights of existing deep learning models. Our unique experiment has given a new observation that will help future researchers working on ensemble learning models. The overall process will help to identify which models are appropriate for making the best decision for detecting skin cancer disease.

The rest of this paper is arranged as follows. Section 2 provides the background needed for the study. The data sources, preprocessing methods, and models used in this work for aggression detection tasks are discussed in Section 3. The simulation results based on the classification algorithms and the comparison using the derived results are analyzed in Section 4. Finally, this paper is summarized in Section 5.

## II. BACKGROUND AND LITERATURE REVIEW

Many investigations have been done on the topic of image classification. We have gone through some of the related papers, which helped us significantly improve our analysis.

Previously, M. Vidya and M. V. Karki [21] showed a machine-learning approach for detecting melanoma skin cancer. Their approach includes five phases: data acquisition, preprocessing of data, segmentation, feature extraction, and classification. They preprocessed the dataset to remove unwanted noises such as artifacts, low contrasts, hairs, veins, skin colors, moles, etc. After that, they used a segmentation process called Geodesic Active Contours (GAC). The features of the shape and edge information are extracted using HOG. Finally, they applied SVM, KNN, Naive Bayes, and Neural Networks to the extracted features. They obtained 97.8 percent accuracy using the SVM classifier and 85 percent specificity using the KNN classifier.

K. Manasa and D. G. V. Murthy [22] used VGG16 and Resnet-50 models for classifying skin cancer disease using malignant and benign images. They used 3297 images to train their models, of which 1497 images belong to the malignant class, and 1800 images belong to the benign class. They got 80 percent accuracy for the VGG16 model and 87 percent accuracy for the Resnet50 model.

M. Hasan et al. [23] used a convolutional neural network to classify cancer images. Their dataset contained benign and malignant classes. Those images are converted into greyscale images for better CPU usage. The preprocessed data are fed into the convolutional neural network. Finally, they evaluated their model using precision, recall, specificity, f1 score, and accuracy. They got 89.5 percent accuracy on the test dataset.

U. B. Ansari and T. Sarode [24] showed an image preprocessing method for detecting skin cancer. Their system is implemented by using Gray Level Co-occurrence Matrix (GLCM) and Support Vector Machine (SVM). GLIM is used to extract features from the image. The features are then fed to the SVM classifier for making the classification result. Before extracting the features, they preprocessed the data by using three methods– Grayscale Conversion, Noise removal, and Image enhancement to reduce unwanted distortions. They preprocessed the images to get the important features from the image. Using their approach, they achieved 95 percent accuracy on the test dataset.

P. G. Brindha et al. [25] showed a comparative study of SVM and CNN for detecting the types of skin cancer. Their image preprocessing method includes reducing the channel of the image by converting the original image into a greyscale image. They used SVM and CNN models to classify skin cancer, where they observed that SVM produced a better result.

T. Saba [26] showed a review on skin cancer analysis. They reviewed the investigations that have been done on the classification of skin cancer. The investigation found that most of the previous research works have been done using the SVM, CNN, and ANN models. Some of the research has been done using the segmentation process, which we have already discussed earlier.

M. A. Kassem et al. [27] proposed a modified GoogleNet model for classifying eight classes of skin lesions. They added more filters to each layer to enhance and reduce noise. They replaced the last three layers in two different ways. The last three layers have been dropped out and replaced with a fully connected layer, a softmax layer, and a classification output layer. That change has been made to increase the probability of the target class. Secondly, the last two layers have been dropped. The original fully connected layer has been kept as same to detect the outliers. They achieved 63 percent accuracy using the original GoogleNet model and 81 percent accuracy using their proposed model, which indicates that their proposed model works better for classification purposes.

D. N. Le et al. [28] used Transfer learning techniques such as pre-trained ResNet50, VGG16, and MobileNet models in combination with focal loss and class weights for classifying the skin cancer. To balance the classes, they used weights in each of the classes. Higher weights were given to the classes with fewer samples, whereas lower weights were assigned to the classes with more samples. Using their approach, they achieved 93 percent average accuracy on the test data.

I. A. Ozkan and M. KOKLU [29] used four different machine learning algorithms such as Artificial Neural Network (ANN), Support Vector Machine (SVM), K-Nearest Neighbors (KNN), and Decision Tree (DT) for classifying melanoma skin cancer. They achieved 92.50 percent accuracy for ANN, 89.50 percent accuracy for SVM, 82.00 percent accuracy for KNN, and 90.00 percent accuracy for DT, which indicates that ANN has a better classification performance.

M. Elgamal [30] used two hybrid approaches to identify skin cancer. At first, they extracted the features using discrete wavelet transformation. After that, the features from the images were reduced using Principal Component Analysis. Finally, the features were fed to the artificial neural network and the k-nearest neighbor classifier to perform the classification. Their approach gave 95 percent accuracy and 97.5 percent accuracy in the two classifiers, respectively.

J. Daghrir et al. [31] classified melanoma skin cancer using the Convolutional Neural Network (CNN), Support Vector Machine (SVM), and K-Nearest Neighbor (KNN) model. Before classifying the images, they segmented the images using Otsu's method. They extracted the features from the segmented images and fed those features to the classifiers. Finally, they combined all three models and made predictions based on the majority votes. Their proposed models gave the best result, which is 88.4 percent accuracy; on the other hand, the KNN model gave 57.3 percent accuracy, the SVM model gave 71.8 percent accuracy, and CNN gave 88.4 percent accuracy.

M. Q. Khan et al. [32] proposed an image processing technique to detect and distinguish melanoma from nevus. At first, they used the Gaussian filter to remove the noise from the images. After that, they used SVM for classifying melanoma and nevus skin cancer. Their proposed methodology achieved almost 96 percent accuracy.

## III. METHODOLOGY

Since the architectures are not developed for multi-classss classification, we propose the generalized architecture for the multi-class classification of skin cancer, shown in Figure 1. At first, all dermoscopic skin cancer images are preprocessed to meet the requirement of models before feeding. The processed images are then fed to the architecture for feature extraction and finetuning. Finally, the input images are divided into seven classes of skin cancer, i.e. Melanocytic Nevi, Melanoma, Benign Keratosis, Actinic Keratosis, Vascular Lesions, Der-matofibroma, and Basal Cell Carcinoma. The classifiers such as InceptionV3, Xception, Densenet, Mobilenet, Resnet-50, CNN, and VGG16 are designed for classifying these seven skin lesion types. Using the weights of the aforementioned models, five different stacking models have been developed. The models are named inceptionv3-inceptionv3, Densenet-mobilenet, inceptionv3-Xception, Resnet50-Vgg16, and stack-six.

Figure 1 illustrates a high-level schematic representation of the classification with existing deep learning models.

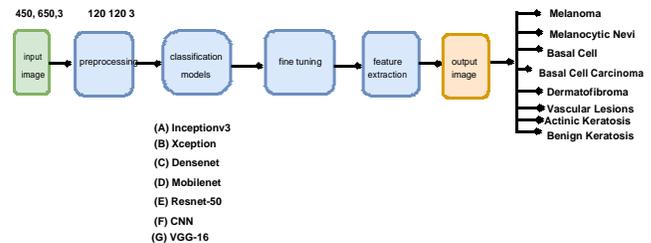

Fig. 1: Process for Multi-class skin cancer classification.

### A. Dataset

The HAM10000 dataset, a large collection of the multi-source dermatoscopic dataset, has been used in this work [33]. It is downloaded from the Kaggle web-site url https://www.kaggle.com/kmader/skin-cancer-mnist-ham10000. The dataset consists of 10,015 skin lesion images of 7 classes. The classes are Melanocytic nevi (6705 images), Melanoma (1113 images), Benign keratosis (1099 images), Basal cell carcinoma (514 images), Actinic keratosis (327 images), Vascular Lesions (142 images), and Dermatofibroma (115 images). All dermoscopic images have a resolution of

600 x 450-pixels, and the channel is three. The images are taken by dermatoscopy instrument. The instrument is a type of magnifier that is used to take pictures of skin lesions.
Figure 2 shows a sample of skin lesion types.

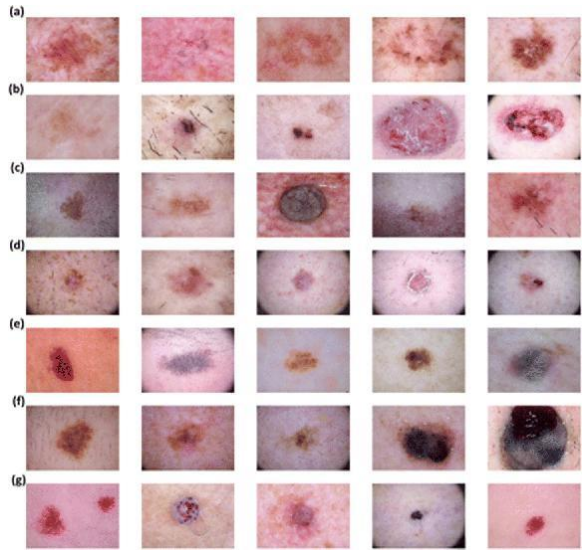

Fig. 2: Sample skin cancer images from HAM10000 dataset (a) Actinic keratosis (b) Basal cellcarcinoma (c) Benign keratosis-like lesions (d) dermatofibroma (e) Melanocytic nevi (f) Melanoma (g) Vascular lesions.

### B. Preprocessing

Raw data is not considered well-prepared inputs for fetching into the deep learning models since the data can have different sizes and noises. Transfer learning models can only work on the image size 224X224 pixels or less than that. Therefore, we have preprocessed the dataset by resizing it from 600 x 450 pixels to 120X120 pixels, and the channel of the images has been kept the same as before. We did not remove the noises, such as hair, discoloration, and other issues, as we wanted our model to learn the noises to perform well on the data with the presence of noise.

### C. Classification Models and Fine Tuning

We perform modifications on the architectures such as Resnet-50, VGG-16, Densenet, Mobilenet, Inceptionv3, and Xception for performing multi-class classification. Deep learning architecture customizations include
1) dense layers with 'relu' activation.
2) dropout layers and softmax layers at the bottom of the architecture.
3) improvement in the parameters' values.
Then, we fine-tune using HAM10000 dataset for classifying skin cancer disease.

1) CNN: A Convolutional Neural Network is a class of neural networks. It is used for image processing and classification. A CNN contains a convolutional layer, a pooling layer, and a fully connected layer. The convolutional layer is the core building block of the CNN model. The convolutional layer performs a dot product between two matrices; one matrix is the kernel, and another is a portion of the input image. The kernel size is spatially smaller than the input image, but the kernel dimension is the same as the input image. A pooling layer is another building block of a CNN. It reduces the spatial size of CNN, which helps reduce the network's parameters and computation. Finally, the fully connected layer is used to provide the final output. The final convolutional neural network output is converted to a flattened one-dimensional input fed to the fully connected layer. The final fully connected layers use the softmax activation function, which gives the probability of input being in a particular class. Finally, we trained the model on 9617 sample images for 30 epochs with a learning rate of 0.0001 and Adam optimizer [34, 35].

2) Inceptionv3: Inceptionv3 was first introduced by Szegedy et al. [36] in 2015. It is a convolutional neural network for analyzing image datasets such as image classi-fication, object localization, and segmentation. The network is an extended version of the GoogleNet model. Inceptionv3 model avoids computational complexity since it concatenates multiple different sizes of convolutional filters into a new filter, which allows the model to reduce the number of parameters to be trained. Therefore, the performance of classification remains well while keeping a fewer number of parameters. Inceptionv3 has become popular among researchers since it can be efficiently trained over a huge dataset. In addition, we have included a dense layer with 'relu' activation, Dropout, and softmax layers with seven outputs at the bottom of the architecture to better fine-tune the model on the dataset. Finally, we fine-tuned the model on 9617 sample images for 30 epochs with a learning rate of 0.0001 and Adam optimizer.

3) Xception: Xception model was first developed by Google researchers named Chollet [37]. Xception is a deep convolutional neural network built with a linear combination of depth-wise separable convolution with residual connections. This novel deep neural network architecture was inspired by the Inception model, where the Inception layers are re-placed with the depth-wise separable convolution. The concept of building the architecture with fully depth-wise separable convolution makes it less complex and more efficient than other deep convolutional neural networks. Moreover, we have included a dense layer with 'Relu' activation, Dropout, and Softmax layers with seven outputs at the bottom of the architecture to fine-tune the dataset model. Finally, we have fine-tuned the model on 9617 images (for 30 epochs) with a learning rate of 0.001 and adam optimizer with a momentum of 0.9.

4) Densenet: A Densely Connected convolutional Network (Densenet) was first proposed by Gao Huang, Zhuang Liu, and their team in 2017 [38]. It is a deep convolutional neural network that uses dense connections between the layers, and each layer connects to each layer in a feed-forward manner. It diminishes the vanishing gradient issue and requires fewer parameters to train the model. So, it has the advantage to use in the computer vision field. Moreover, we have included a dense layer with 'Relu' activation, Dropout, and Softmax layers with seven outputs at the bottom of the architecture to fine-tune the dataset model. Finally, we have fine-tuned the model on 9617 images (for 30 epochs) with a learning rate of 0.001 and adam optimizer with a momentum of 0.9.

5) Mobilenet: Andrew G. Howard et al. [39], first intro-duced MobileNet architecture. Among the deep neural net-works, Mobilenet is a lightweight model which is appropriate for reducing computational cost and time. To reduce the model size and computation, MobileNet uses depthwise separable convolutions instead of standard convolutions. Depthwise sep-arable convolution is a factorized convolution that factorizes standard convolution into two convolutions: a depthwise con-volution and a pointwise convolution. Pointwise convolution is a 1* 1-dimensional convolution. Depthwise convolution aims to filter, whereas the purpose of pointwise convolution is to combine. The sum of depthwise convolution and pointwise convolution is the operation of depthwise separable convolu-tion. The architecture of MobileNet is built with separable con-volutions except for the first layer. The first layer is built with a complete convolutional layer. In addition, we have included a dense layer with 'Relu' activation, Dropout, and Softmax layers with seven outputs at the bottom of the architecture to better fine-tune the model on the dataset. Finally, we fine-tuned the model on 9617 sample images for 30 epochs with a learning rate of 0.0001 and Adam optimizer.

6) Resnet-50: A residual Neural Network( ResNet) is a kind of Artificial Neural Network(ANN) that forms a net-work by stacking residual blocks on top of each. Resnet has many variants, and the most popular networks are ResNet-34, ResNet-50, and ResNet-101. Each variant follows the same concept of Resnet; the only difference is in the number of layers. Resnet-50 works on 50 neural network layers. Large layers are useful for solving complex problems as each of the layers deals with a unique task. But the problem with the deeper network is it shows a degradation issue. Usually, the degradation problem causes either by the initialization of the network, by the optimization function, or by the problem of vanishing or exploding gradients. The Resnet model aims to avoid such issues. The strength of the Resnet model is the skip connection, which lies at the core of the residual blocks and is responsible for avoiding the degradation problem. Skip connections work in two ways. Firstly, they extenuate the vanishing gradient issue by creating an alternate shortcut for passing the gradient. Secondly, they allow the model to learn an identity function that ensures that the higher layers work almost similarly to the lower ones. Since the model is trained on more than a million images, it can classify the small dataset accurately. In addition, we have included a dense layer with 'Relu' activation, Dropout, and Softmax layers with seven outputs at the bottom of the architecture to better fine-tune the model on the dataset. Finally, we fine-tuned the model on 9617 sample images for 30 epochs with a learning rate of 0.0001 and Adam optimizer [40].

7) VGG-16: VGG-16 is the state-of-the-art deep neural network for analyzing image input. The model was used to win ILSVR (Imagenet) competition in 2014 and is considered one of the excellent vision model architectures to date. It is a large network and contains 138 million parameters. The number 16 of the VGG16 network refers to 16 layers with weights. The unique thing about vgg16 is that it maintains a convolution layer of 3x3 filter with a stride 1, and the same padding and max pool layer of a 2x2 filter of stride 2 , throughout the whole architecture. Finally, the model ends with 2 Fully Connected Layers followed by a softmax for output. In addition, we have included a dense layer with 'Relu' activation, Dropout, and Softmax layers with seven outputs at the bottom of the architecture to better fine-tune the model on the dataset. Finally, we fine-tuned the model on 9617 sample images for 30 epochs with a learning rate of 0.0001 and Adam optimizer [41].

D. Proposed stacking models

1) Inceptionv3-Inceptionv3: Inceptionv3-Inceptionv3 stack-ing model is developed using the weights derived from two inceptionv3 models. The models are trained with the same input but individually. The weights are saved in a folder for further process. The weights are fed to a decision tree model for final prediction. The decision tree model worked as a meta-model. Meta model is a fast learner since it predicts the input, which is already the predictions from the classifiers.

2) Densenet-Mobilenet: Densenet-Mobilenet stacking model is developed using the weights derived from one Densenet and one Mobilenet model. The models are trained with the same input but individually. The weights are saved in a folder for further process. The weights are fed to a decision tree model for final prediction. The decision tree model worked as a meta-model. Meta model is a fast learner since it predicts the input, which is already the predictions from the classifiers.

3) Inceptionv3-Xception: Inceptionv3-Xception stacking model is developed using the weights derived from one Inceptionv3 and one Xception model. The models are trained with the same input but individually. The weights are saved in a folder for further process. The weights are fed to a decision tree model for final prediction. The decision tree model worked as a meta-model. Meta model is a fast learner since it predicts the input, which is already the predictions from the classifiers.

4) Resnet50-Vgg16: Resnet50-Vgg16 stacking model is developed using the weights derived from one Resnet50 and one vgg16 model.The models are trained with the same input but individually. The weights are saved in a folder for further process. The weights are fed to a decision tree model for final prediction. The decision tree model worked as a meta-model. Meta model is a fast learner since it predicts the input, which is already the predictions from the classifiers.

5) Stack-Six: Stack-six stacking model is developed using the weights derived from six models such as Resnet-50, VGG-

16, Densenet, Mobilenet, Inceptionv3, and Xception. The models are trained with the same input but individually. The weights are saved in a folder for further process. The weights are fed to a decision tree model for final prediction. The decision tree model worked as a meta-model. Meta model is a fast learner since it predicts the input, which is already the predictions from the classifiers.

E. Evaluation Metrics

Evaluating a model's performance is necessary since it gives an idea of how close the model's predicted outputs are to the corresponding expected outputs. The evaluation metrics are used to evaluate a model's performance. However, the evaluation metrics differ with the types of models. The types of models are classification and regression. Regression refers to the problem that involves predicting a numeric value. Classification refers to the problem that involves predicting a discrete value. The regression problem uses the error metric for evaluating the models. Unlike the regression problem model, the classification problem uses the accuracy metric for evaluation. Since our motive is to classify the cancerous cell, we used accuracy, f1 score, precision, and Recall for our evaluation metric [42–45].

Precision : When the model predicts positive, it should be specified that how much the positive values are correct. Precision is used when the False Positives are high. For skin cancer classification, if the model gives low precision, then many non-cancerous images will be detected as cancerous; for high precision, it will ignore the False positive values by learning with false alarms. The precision can be calculated as follows:

$$Precision = \frac{TP}{TP + FP} \quad (1)$$

Here TP refers to True Positive values and FP refers to False Positive values.

Recall : The metric recall is the opposite of Precision. The precision is used when the false negatives (FN) are high. In the aggressive detection classification problem, if the model gives low recall, then many cancerous cells will be said as non-cancerous cells; for high recall, it will ignore the false negative values by learning with false alarms. The recall can be calculated as follows:

$$Recall = \frac{TP}{TP + FP} \quad (2)$$

F1 score: F1 score combines precision and recall and provides an overall accuracy measurement of the model. The value of the F1 score lies between 1 and 0. If the predicted value matches with the expected value, then the f1 score gives 1, and if none of the values matches with the expected value, it gives 0. The F1 score can be calculated as follows:

$$F1score = \frac{2 \cdot precision \cdot recall}{precision + recall} \quad (3)$$

Accuracy : Accuracy determines how close the predicted output is to the actual value.

$$Accuracy = \frac{TP + TN}{TP + TN + FP + FN} \quad (4)$$

here, TN refers to True Negative and FN refers to False Negative.

IV. RESULTS AND DISCUSSIONS

A. Results of existing deep learning models

The model's accuracy is derived from the validation data containing 1103 images of seven classes. We have used TensorFlow and Keras library for implementing the Neural Network models. Model training is done on the Google Colab using GPU runtime. We have evaluated the performance of seven different models such as Inceptionv3, Xception, Densenet, Mobilenet, Resnet-50, CNN, and VGG-16 for the classification of skin cancer among seven classes: Melanocytic nevi, Melanoma, Benign keratosis, Basal cell carcinoma, Actinic keratosis, Vascular Lesions, and Dermatofibroma using performance metrics such as precision, recall, f1-score, and accuracy. The categorical accuracy for Inceptionv3, Xception, Densenet, Mobilenet, Resnet-50, CNN, and VGG-16 are 90 percent, 88 percent, 88 percent, 87 percent, 82 percent, 77 percent, and 73 percent, respectively. The inceptionv3 model is provided the best result among all the models. The weighted average of precision, recall, and F1-score for InceptionV3 is 91 percent, 90 percent, and 90 percent, respectively. Similarly, the weighted averages of precision, recall, and F1-score for Xception, Densenet, Mobilenet, Resnet-50, CNN, and VGG-1 are also evaluated, which are shown in Table 1.

TABLE-1 : Results of Inceptionv3, Xception, Densenet, Mobilenet, Resnet-50, CNN, and VGG-1 models

| Models | Accuracy | Precision | Recall | F1-Score |
|---|---|---|---|---|
| Inceptionv3 | 0.90 | 0.90 | 0.90 | 0.90 |
| Xception | 0.88 | 0.88 | 0.88 | 0.87 |
| Densenet | 0.88 | 0.88 | 0.88 | 0.87 |
| Mobilenet | 0.87 | 0.88 | 0.87 | 0.86 |
| Resenet-50 | 0.82 | 0.80 | 0.82 | 0.81 |
| CNN | 0.77 | 0.73 | 0.77 | 0.73 |
| VGG-16 | 0.73 | 0.71 | 0.73 | 0.71 |

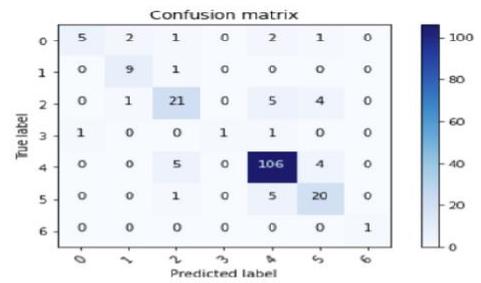

(a)

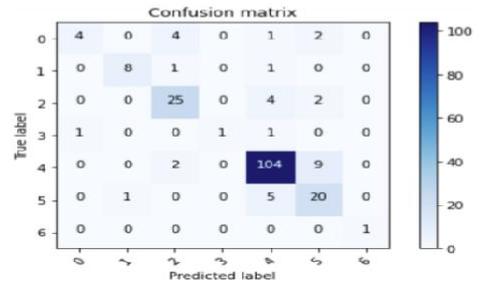

(b)

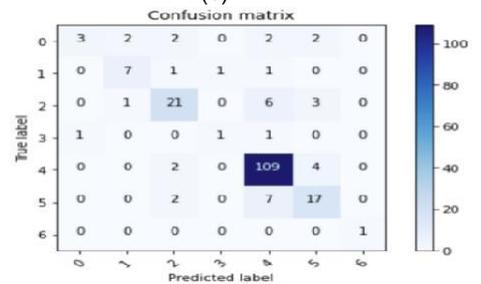

(c)

For each of the seven models, The training-validation accuracy and training-validation loss curves are represented in Figure 3. For all models, the training accuracy is higher than the validation accuracy, or the training loss is lower than the validation loss from the first epochs. One possible observation can be adding the Dropout layer in the architecture during fine-tuning of the model since it makes a pretrained model less prone to over-fitting. These Dropout layers disable the neurons during training to reduce the complexity of the model.

In Figure 3, the confusion matrix shows the number of True positive and False negative results has been predicted by each of the model.

B. Results of Developed Stacking Models

Using the weights of the seven aforementioned classifiers, we develop five stacking ensemble models such as Inceptionv3-Inceptionv3, Densenet-Mobilenet, Inceptionv3-Xception, Resnet-50-vgg16, and stack-six. We evaluate the models performance with the same test dataset which is used for the seven aforementioned models. The weighted averages of precision, recall, and F1-score for Inceptionv3-Inceptionv3, Densenet-Mobilenet, Inceptionv3-Xception, Resnet-50-vgg16, and stack-six are also evaluated, shown in Table 2.

Table 2: Results of inceptionv3-inceptionv3, Densenet-mobilenet, inceptionv3-Xception, Resnet50-Vgg16, and stack-six model.

| Models | Accuracy | Precision | Recall | F1-Score |
|---|---|---|---|---|
| Inceptionv3-Inceptionv3 | 0.78 | 0.79 | 0.78 | 0.78 |
| Densenet-Mobilenet | 0.78 | 0.77 | 0.78 | 0.77 |
| Inceptionv3-Xception | 0.75 | 0.74 | 0.75 | 0.74 |
| Resnet-50-vgg16 | 0.70 | 0.72 | 0.70 | 0.71 |
| stack-six | 0.78 | 0.80 | 0.78 | 0.77 |

The result shows that the stacking ensemble model provides the highest accuracy of 78 percent, which is lesser than the existing deep learning model's performance. Since the stacking ensemble models are less prone to show bias result, it can be the reason why it shows poor result and does not vary the result within the stacking models. The lowest accuracy is 70 and the highest accuracy is 78. The difference is very less. Whereas, The single models accuracy varies largely from one model to another. The lowest accuracy is 73 and the highest accuracy is 90. So, the models could have tendency of showing bias result for a particular dataset.

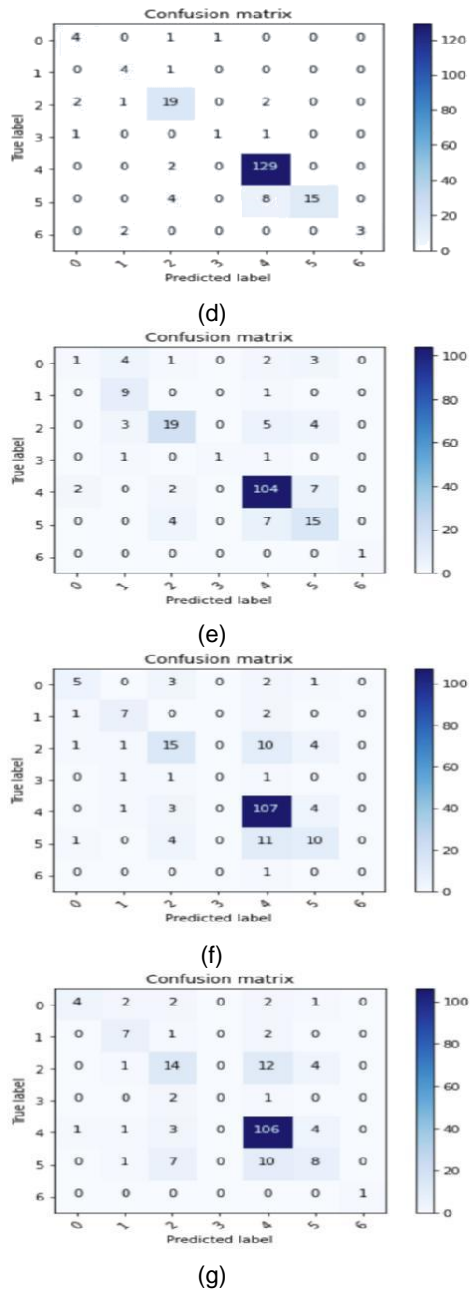

Fig. 3: Confusion matrix derived from a) inceptionv3 (b) Xception (c) Densenet (d) Mobilenet (e) Resenet (f) CNN (g) Vgg16 model using HAM1000 skin cancer dataset.

We stack a model with six models and named it stack-six model. The stack-six model gives 78 percent accuracy and did not improve much even though we have increased the number of weights for stacking. Therefore, it is clear that the stacking model's performance becomes saturated for a particular accuracy range.

## V. CONCLUSION

Since the death rate due to skin cancer increases day by day, it is necessary to address this global public health issue. The outstanding performance of deep convolution models on image datasets can be utilized for skin cancer detection. However, using the deep learning models, a different problem has a different process to solve. Previously, several investigations have been done on classifying skin cancer detection. To the best of our knowledge, none of the work has shown the comparative analysis of multiple skin cancer classes using deep convolutional neural networks. Our work will help the medical sector distinguish skin cancer from different skin lesions accurately. Seven classes of skin lesions have been classified using Resnet-50, VGG-16, Densenet, Mobilenet, Inceptionv3, Xception, and CNN. Finally, the performance of the models is evaluated using evaluation metrics such as precision, recall, f1-score, and accuracy. Among all the models, Inceptionv3 provides the best result, which is 90 per-cent accuracy. Furthermore, we have developed five stacking ensemble models such as inceptionv3-inceptionv3, Densenet-Mobilenet, inceptionv3-Xception, Resnet50-Vgg16, and stack-six using the weights of the aforementioned models for observing how the stacked models perform individually on the same dataset. We have found that stacking models give the highest accuracy of 78 percent, which is lesser than the performance of the existing models. Since single models have a tendency to show biased results, it can be one reason why the accuracy widely varies from one model to another. Therefore, our experiment shows that the stacking model may give a higher lower accuracy than the existing model since the model does not show biased results. Our work will give a clear observation of stacking ensemble models to future researchers for further investigations on the skin cancer dataset as well as the ensemble learning models.